\documentclass{article}

\usepackage[final]{corl_2024} 

\usepackage{amssymb,amsmath}
\usepackage{cite} 
\usepackage{booktabs}
\usepackage[font=footnotesize,labelformat=simple]{subcaption}
\usepackage[font=footnotesize]{caption}
\usepackage{attachfile}

\usepackage{thmtools,thm-restate}
\usepackage[noend, ruled, linesnumbered]{algorithm2e}
\usepackage[noend]{algpseudocode}
\usepackage{multirow}
\usepackage{tabulary}
\usepackage{graphicx}
\usepackage{subcaption}

\usepackage{hyperref}
\hypersetup{bookmarksopen,bookmarksnumbered,
pdfpagemode=UseOutlines,
colorlinks=true,
linkcolor=blue,
anchorcolor=blue,
citecolor=blue,
filecolor=blue,
menucolor=blue,
urlcolor=blue
}

\graphicspath{{./figs/}}
\usepackage{xspace}

\usepackage{enumitem}
\newlist{hypothesis}{enumerate}{1}
\setlist[hypothesis]{resume, label=\textbf{H\arabic*}, labelindent=\parindent, leftmargin=*}
\usepackage{xcolor}
\definecolor{orange}{rgb}{1,0.5,0}
\definecolor{internationalorange}{rgb}{1.0, 0.31, 0.0}

\newcommand{\xxnote}[3]{}
\ifx\hidenotes\undefined
  \usepackage{color}
  \renewcommand{\xxnote}[3]{\color{#2}{#1: #3}}
\fi

\title{Parental Guidance: Efficient Lifelong Learning through Evolutionary Distillation}
\author{
  Octi Zhang\textsuperscript{†},
  Quanquan Peng\textsuperscript{‡}, 
  Rosario Scalise\textsuperscript{†},
  Byron Boots\textsuperscript{†}
  \\
  \textsuperscript{†}Paul G Allen School, University of Washington \\
  \textsuperscript{‡}Shanghai Jiao Tong University \\
}

\begin{document}
\maketitle

\vspace{-0.7cm}
\begin{abstract}


Developing robotic agents that can perform well in diverse environments while showing a variety of behaviors is a key challenge in AI and robotics. Traditional reinforcement learning (RL) methods often create agents that specialize in narrow tasks, limiting their adaptability and diversity. To overcome this, we propose a preliminary, evolution-inspired framework that includes a reproduction module, similar to natural species reproduction, balancing diversity and specialization. By integrating RL, imitation learning (IL), and a coevolutionary agent-terrain curriculum, our system evolves agents continuously through complex tasks. This approach promotes adaptability, inheritance of useful traits, and continual learning. Agents not only refine inherited skills but also surpass their predecessors. Our initial experiments show that this method improves exploration efficiency and supports open-ended learning, offering a scalable solution where sparse reward coupled with diverse terrain environments induces a multi-task setting.


\end{abstract}


\section{Introduction}
\label{sec:intro}

Developing robotic agents that can generalize across diverse environments while continually evolving their behaviors is a core challenge in AI and robotics. The difficulties lie in solving increasingly complex tasks and ensuring agents can continue learning without converging on narrow, specialized solutions. Quality Diversity (QD) \citep{wang2019paired, lehman2011evolving} methods effectively foster diversity but often rely on trial and error, where the path to a final solution can be convoluted, leading to inefficiencies and uncertainty.

Our approach draws inspiration from nature's inheritance process, where offspring not only receive but also build upon the knowledge of their predecessors. Similarly, our agents inherit distilled behaviors from previous generations, allowing them to adapt and continue learning efficiently, eventually surpassing their predecessors. This natural knowledge transfer reduces randomness, guiding exploration toward more meaningful learning without manual intervention like reward shaping or task descriptors.

What sets our method apart is that it offers a straightforward, evolution-inspired way to consolidate and progress, avoiding the need for manually defined styles or gradient editing \citep{yu2020gradient, kirkpatrick2017overcoming} to prevent forgetting. The agent’s ability to retain and refine skills is driven by a blend of IL and RL, naturally passing down essential behaviors while implicitly discarding inferior ones.

We introduce \textbf{P}arental \textbf{G}uidance (\textbf{PG-1}) which makes the following contributions:
\begin{enumerate}

    \item \textbf{Distributed Evolution Framework}: We propose a framework that distributes the evolution process across multiple compute instances, efficiently scheduling and analyzing evolution.

    \item \textbf{Integration of Learning Paradigms}: We integrate RL to find locally optimal policies, automatic curriculum to continually challenge the learners, and IL to pass down specialists' optimal solutions via distillation within the aforementioned evolutionary framework.

    \item \textbf{Preliminary Evaluation of Method Efficacy}:
     We evaluate our framework using behavior cloning (an IL method) and RL as baselines. 
     We compare a number of integration strategies, and we provide guidance on how to best transition between IL and RL to maximize the benefits of our framework.

\end{enumerate}


\section{Related Works}
\label{sec:related works}

\textbf{Quality Diversity and Evolutionary Approaches}
 QD algorithms, combined with evolutionary methods \citep{wang2019paired, mouret2015illuminating, inproceedings}, promote behavioral diversity by exploring a range of solutions optimized for different niches. These approaches have shown success in control tasks \citep{petrenko2023dexpbt} by expanding skill sets within a parent's nearby space, improving exploration. However, single-lineage inheritance can struggle when tasks require joint capabilities discovered in distant regions. In nature, species accelerate exploration by learning from each other, and passing knowledge across generations. Mammals, known for their intelligence and complex social behavior, all reproduce sexually, an intriguing biological parallel to how we approach knowledge sharing in algorithms. Through mechanisms like knowledge distillation or reproduction, agents can share behaviors, develop generalizable skills, and reduce dependence on long inheritance cycles.

\textbf{RL, IL in Context of Continual Learning}
RL \citep{mnih2013playing, silver2014deterministic} has been shown to excel at specialized tasks but struggles to generalize beyond predefined hand-crafted rewards. IL effectively replicates expert behaviors but often lacks adaptability and improvement. The continual learning community recognizes this divergence and has begun to explore bridging RL and IL to mirror the natural process of parent-child learning \citep{hadsell2020embracing}, where a child inherits knowledge from both parents before becoming independent and excelling in its own unique way. A successful outcome in this context implies that the student policy’s growth stems from both parental skills and its own continuous exploration. Our prelimiary experimentation is designed to validate this intuition. While previous works have focused on distillation techniques \citep{rusu2015policy, hinton2015distilling} that ensure knowledge transfer, the transition from IL to RL remains underexplored in the context of coninual learning. We explore the IL-to-RL transition in a natural evolution framework, discussing its role with respect to the wholistic evolutionary framework.


\section{Methodology}
\label{sec:methodology}
\label{sec:methodology}

\subsection{Distributed Evolution with a Central Orchestrator}

At the core of our framework is a central scheduler that manages an evolutionary tree, represented as a Directed Acyclic Graph (DAG). The scheduler selects the next species to train based on the current state, governed by an evolutionary Markov Decision Process (MDP). Each job runs on a dockerized compute instance of Nvidia Isaac Lab \citep{mittal2023orbit}, which requests tasks from the scheduler. The scheduler handles the outer-level evolutionary progression and data management, while the compute instances focus on executing the inner-level training.

\subsection{Continual Learning via Behavioral Inheritance and Reinforcement Learning}

Behavioral inheritance refers to the process where offspring agents inherit traits and behaviors from their parent species. This inheritance is facilitated through the DAgger~\citep{ross2011reduction}, enabling agents to blend behaviors across multiple parent species. The process includes two key phases, weighted by a BC decay rate $\lambda \in (0, 1)$. During behavior distillation, offspring are trained on aggregated datasets derived from the experiences of both parents, allowing them to inherit and combine a diverse range of behaviors. As training progresses, RL progressively takes over to further refine and enhance their skills. We employed Proximal Policy Optimization (PPO) ~\citep{schulman2017proximal}, though this approach is flexible and can accommodate other RL algorithmic variants. This phase enables agents to move beyond inherited behaviors, cultivating their unique capabilities while building on their parents' foundation. For further details on the learning process, see Algorithm.~\ref{alg:learn_with_expert}.

\begin{figure}[!t]
\vspace{-0.7cm}
\centering

\includegraphics[width=1.0\linewidth , trim={0 0 0 0}, clip]{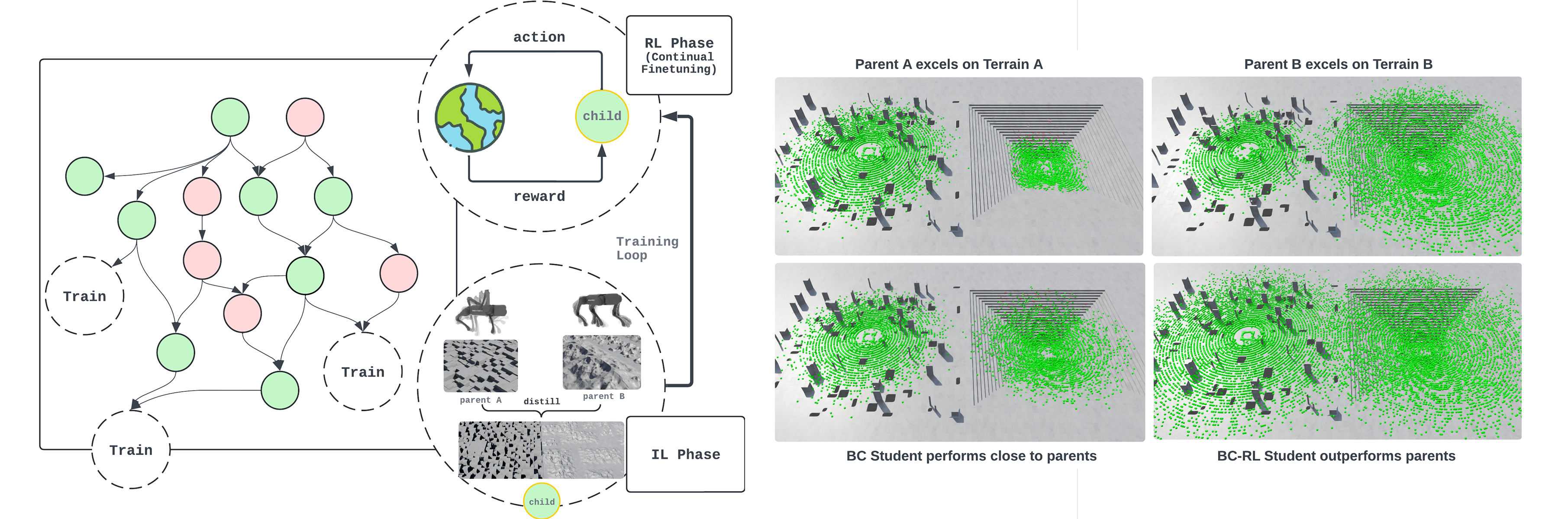}

\caption{\textbf{Left}: Setup of Holistic Evolutionary Framework, an evolution scheduler that maintains a phylogenetic tree (a directed acyclic graph(DAG)). Each training cycle can be modularized and dispatched as a standalone process to an arbitrarily scalable number of compute nodes. Each node performs BC and RL and attaches the new child (the future parent) back to the phylogenetic tree. \textbf{Right}: Comparison of effect of BC-RL process. Each parent is a specialist in its own niche terrain and performs worse in other terrains. Each green dot represents treats successfully fetched by the agent. We let the agent run until the distance increment stops, indicating the agent skill has saturated. Distillation enables agents to do almost as well as the parent in both terrains, but BC-RL is able to exceed the performance of both parents on the union of tasks.}
\label{fig:best}

\vspace{-0.4cm}
\end{figure}





\subsection{Training Setup}

We set up training with a locomotion task, where the goal of the agent is to control its joints to fetch a treat. The treat in the environment is defined by its position \( x_{\text{treat}} \), which is initialized near the agent’s starting position \( x_{\text{start}} \). The treat’s position remains the same until the agent successfully fetches it. Once fetched, the treat spawns farther away. Specifically, the spawning radius of the treat is updated based on the agent’s success at fetching the treat:

\vspace{-0.3cm}

\begin{equation}
x^{\text{episode} + 1}_{\text{treat}} = 
\begin{cases}
x^{\text{episode}}_{\text{treat}} + \Delta x_{\text{treat}}, & \text{if } \|x_{\text{agent}, t} - x^{\text{episode}}_{\text{treat}}\| \leq \epsilon, \\
x^{\text{episode}}_{\text{treat}} - \Delta x_{\text{treat}}, & \text{otherwise}
\end{cases}
\label{eq:terrain}
\end{equation}
 \( \Delta x_{\text{treat}} > 0 \) is the increment in the treat’s position, \( x_{\text{agent}, t} \) is the agent’s position at time step \( t \), and \( \epsilon \) is a small threshold indicating when the agent has successfully fetched the treat.

The overall objective of the student policy, $\pi_s$, maximizes the convex combination of  $\mathcal{L}_{\text{BC}}$ and  $\mathcal{L}_{\text{RL}}$ which seeks to balance both sub-objectives. The surrogate loss $\mathcal{L}_{\text{surrogate}}$ and $\mathcal{L}_{\text{value}}$ are defined as in PPO~\citep{schulman2017proximal}:

\vspace{-0.4cm}
\begin{equation}
    \mathcal{L}_{\text{BC}} = \frac{1}{T}\sum_{t=0}^T \left( \mathcal{L}_{\text{BC}}^{\mu,t} + \mathcal{L}_{\text{BC}}^{\sigma,t} \right)
\label{eq:bcloss}
\end{equation}
\vspace{-0.2cm}
\begin{equation}
    \mathcal{L}_{\text{RL}} = \mathcal{L}_{\text{surrogate}} - c \cdot 
 \textit{KL}[\pi_{s\_old}{(\cdot | s_t)}, \pi_{s}(\cdot | s_t)] + \beta \cdot \mathcal{L}_{value}
\label{eq:rlloss}
\end{equation}
\vspace{-0.2cm}
\begin{equation}
    \mathcal{L} = \lambda^i \cdot \mathcal{L}_{\text{BC}} + (1 - \lambda^i) \cdot \mathcal{L}_{\text{RL}}
\label{eq:combined}
\end{equation}

\begin{algorithm}[h]
\SetAlgoLined
\caption{Method Overview}
\label{alg:learn_with_expert}
\KwIn{Number of iterations $N$, Expert policies $\{\pi_E^t\}$, Student policies $\pi_{S}$ }
Initialize $\pi_{S}$, $\lambda$; \Comment{$\lambda$ is bc weight decay rate}\;
\For{i $=1$ \KwTo $N$}{
    \For{step $=1$ \KwTo num\_steps\_per\_env}{
        Use policy $\pi_{S}$ to sample rollouts $\mathcal D:=\{(obs, a_s, reward)\}$ \;
        \For{each terrain $t$}{
            Obtain expert $\pi_E^t$ actions mean $\mu_E^t$ and std $\sigma_E^t$ from $\pi_E^t$\;
        }
    }
    \For{each $d$ in $\mathcal D$}{
        
        Compute total loss $\mathcal{L}$ and update $\pi_{S}$ based on Eqn.~\ref{eq:combined}
    }
}
\end{algorithm}

\vspace{-0.2cm}
\section{Experimental Results}
\label{sec:results}




\begin{figure*}[!htb]
\centering
\scriptsize{\textbf{Terrain Score}} \hspace{3cm} \scriptsize{\textbf{BC Loss}} \hspace{3cm} \scriptsize{\textbf{Reward}}

\vspace{0.2cm} 

\begin{minipage}{0.05\textwidth} 
    \scriptsize{\textbf{child 1}}
\end{minipage}
\begin{minipage}{0.28\textwidth}
    \centering
    \includegraphics[width=\linewidth, trim={0 0 0 200}, clip]{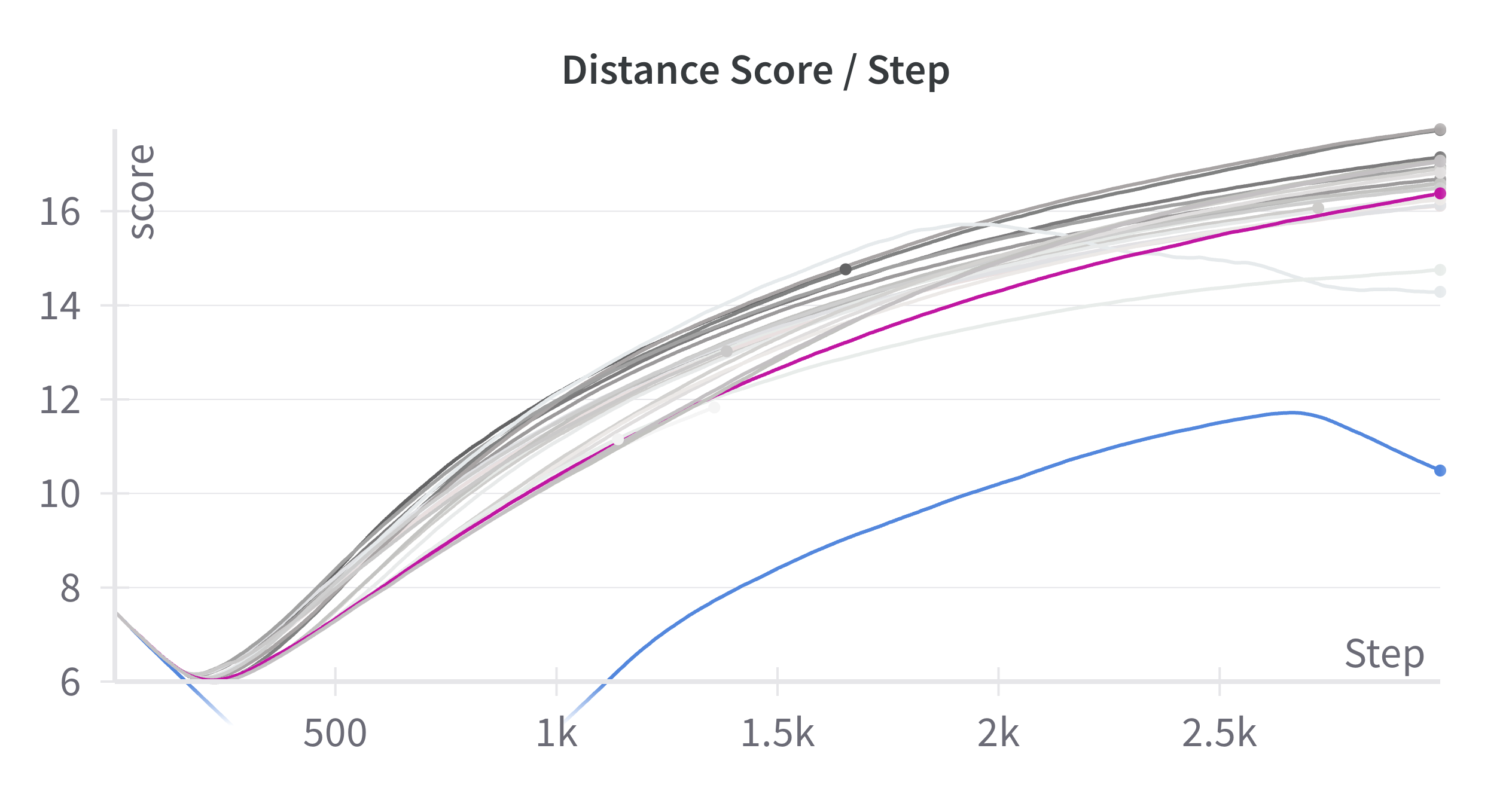}
    \caption*{}
\end{minipage}
\begin{minipage}{0.28\textwidth}
    \centering
    \includegraphics[width=\linewidth, trim={0 0 0 200}, clip]{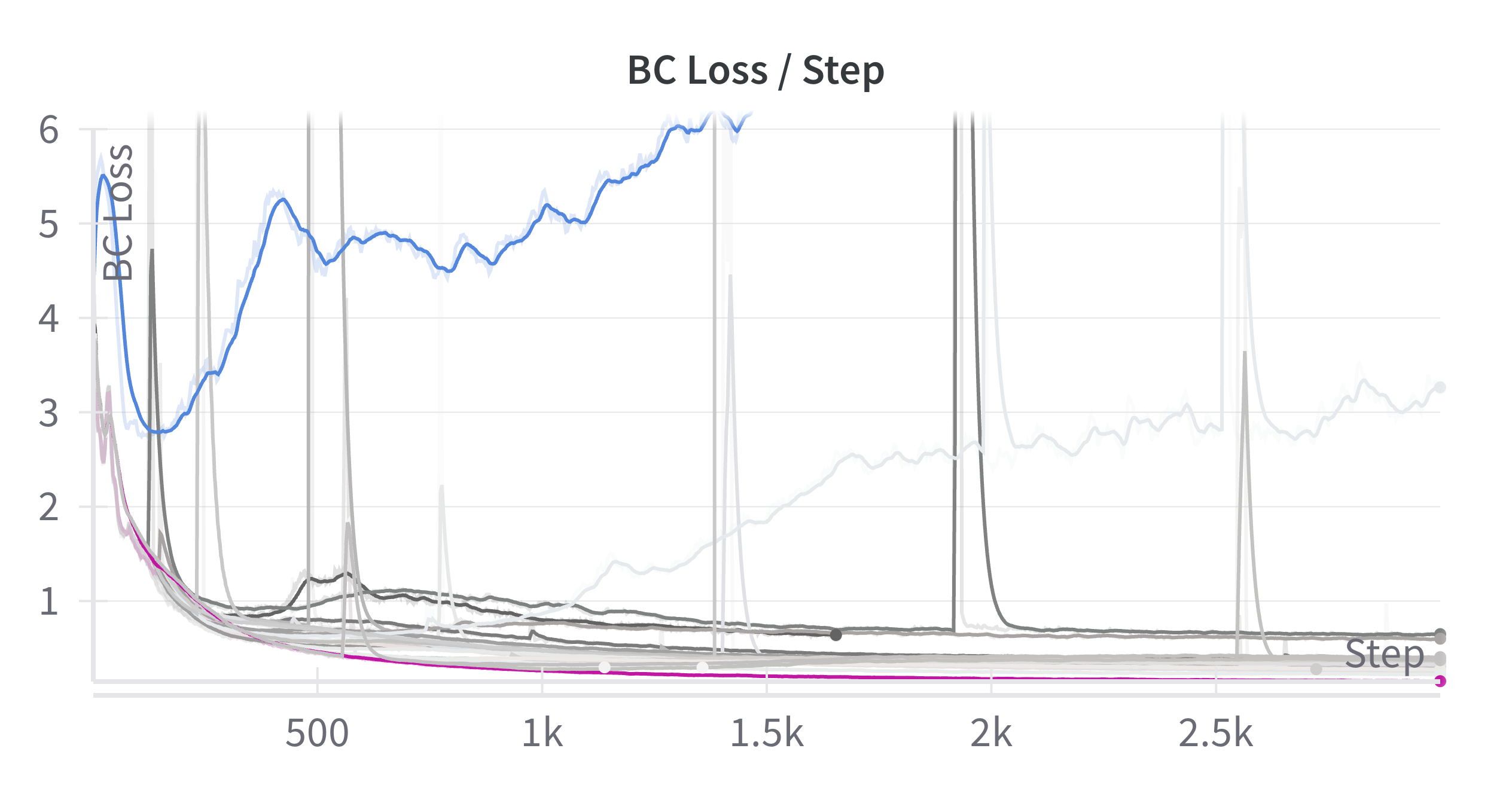}
    \caption*{}
\end{minipage}
\begin{minipage}{0.28\textwidth}
    \centering
    \includegraphics[width=\linewidth, trim={0 0 0 200}, clip]{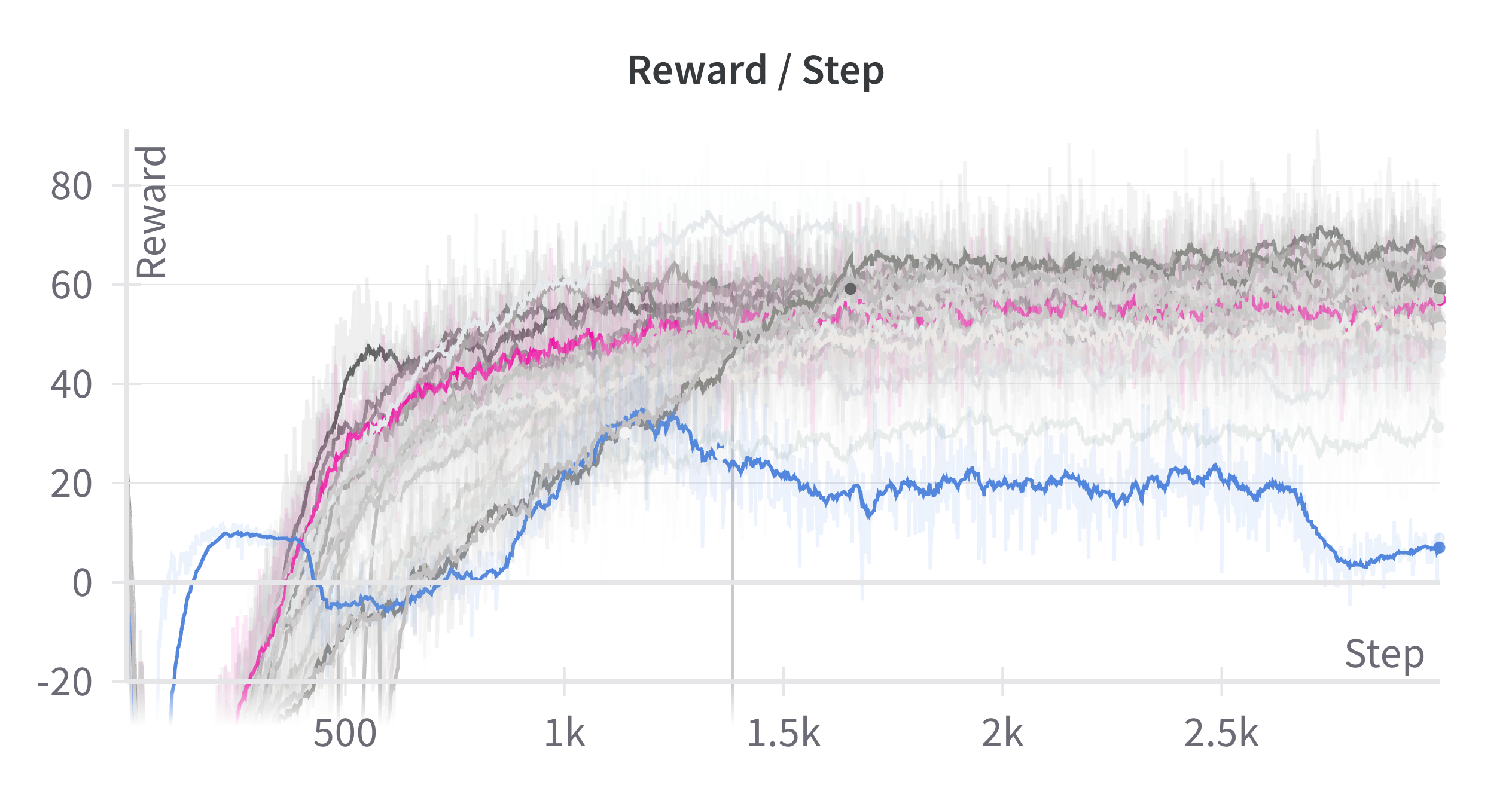}
    \caption*{}
\end{minipage}

\vspace{-0.5cm} 

\begin{minipage}{0.05\textwidth} 
    \scriptsize{\textbf{child 2}}
\end{minipage}
\begin{minipage}{0.28\textwidth}
    \centering
    \includegraphics[width=\linewidth, trim={0 0 0 200}, clip]{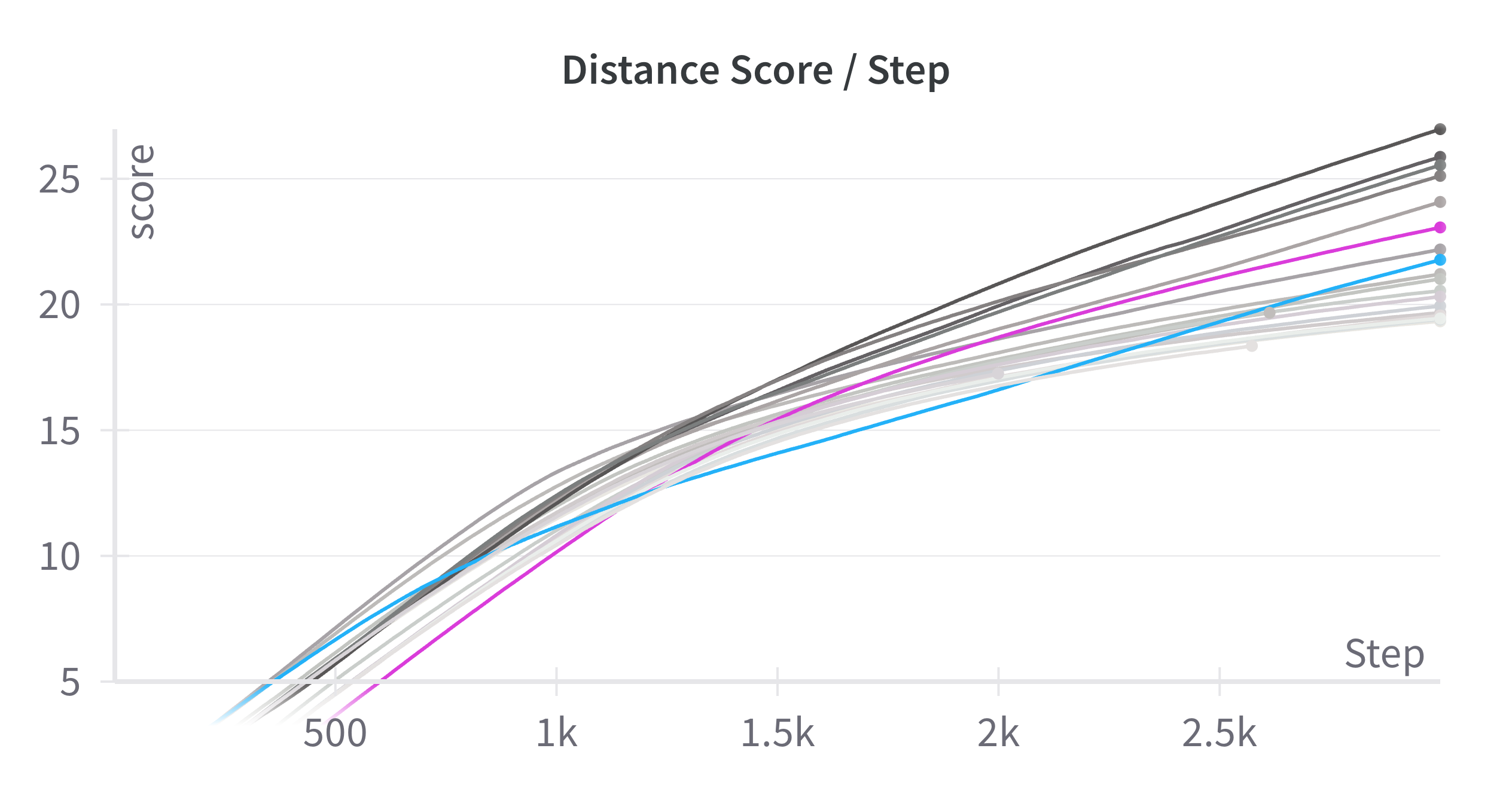}
    \caption*{}
\end{minipage}
\begin{minipage}{0.28\textwidth}
    \centering
    \includegraphics[width=\linewidth, trim={0 0 0 200}, clip]{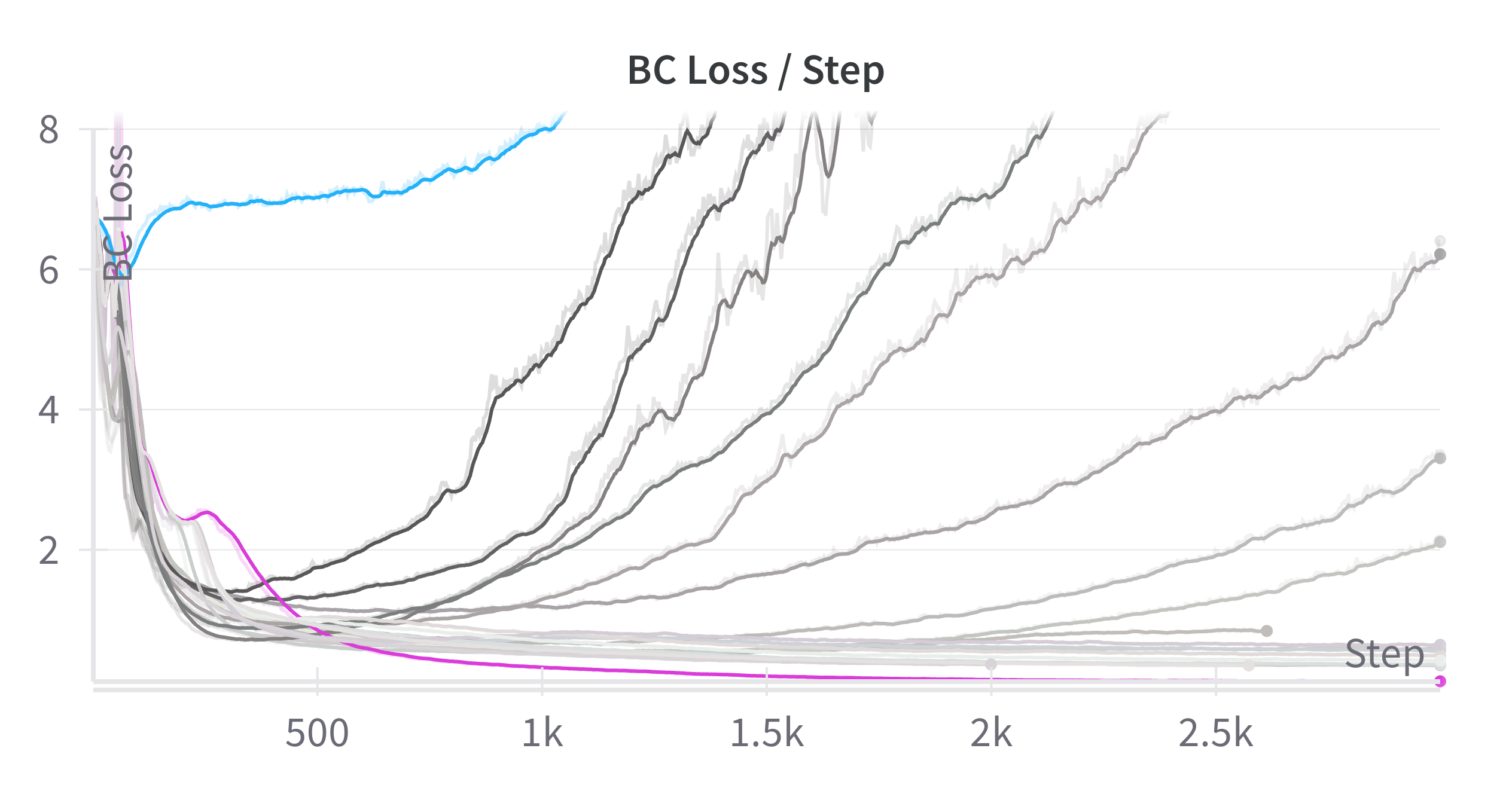}
    \caption*{}
\end{minipage}
\begin{minipage}{0.28\textwidth}
    \centering
    \includegraphics[width=\linewidth, trim={0 0 0 200}, clip]{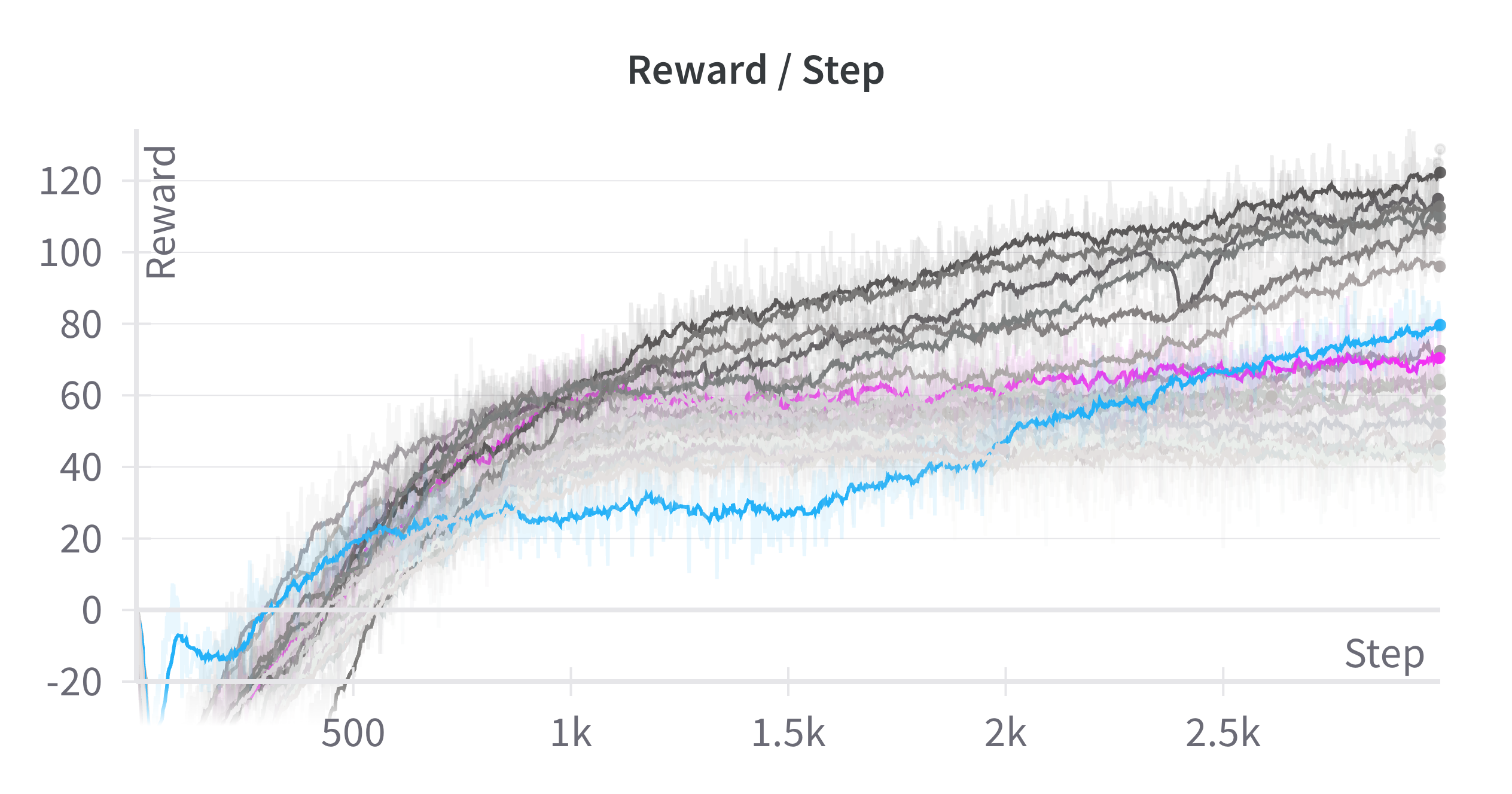}
    \caption*{}
\end{minipage}

\vspace{-0.5cm} 

\begin{minipage}{0.05\textwidth} 
    \scriptsize{\textbf{child 3}}
\end{minipage}
\begin{minipage}{0.28\textwidth}
    \centering
    \includegraphics[width=\linewidth, trim={0 0 0 200}, clip]{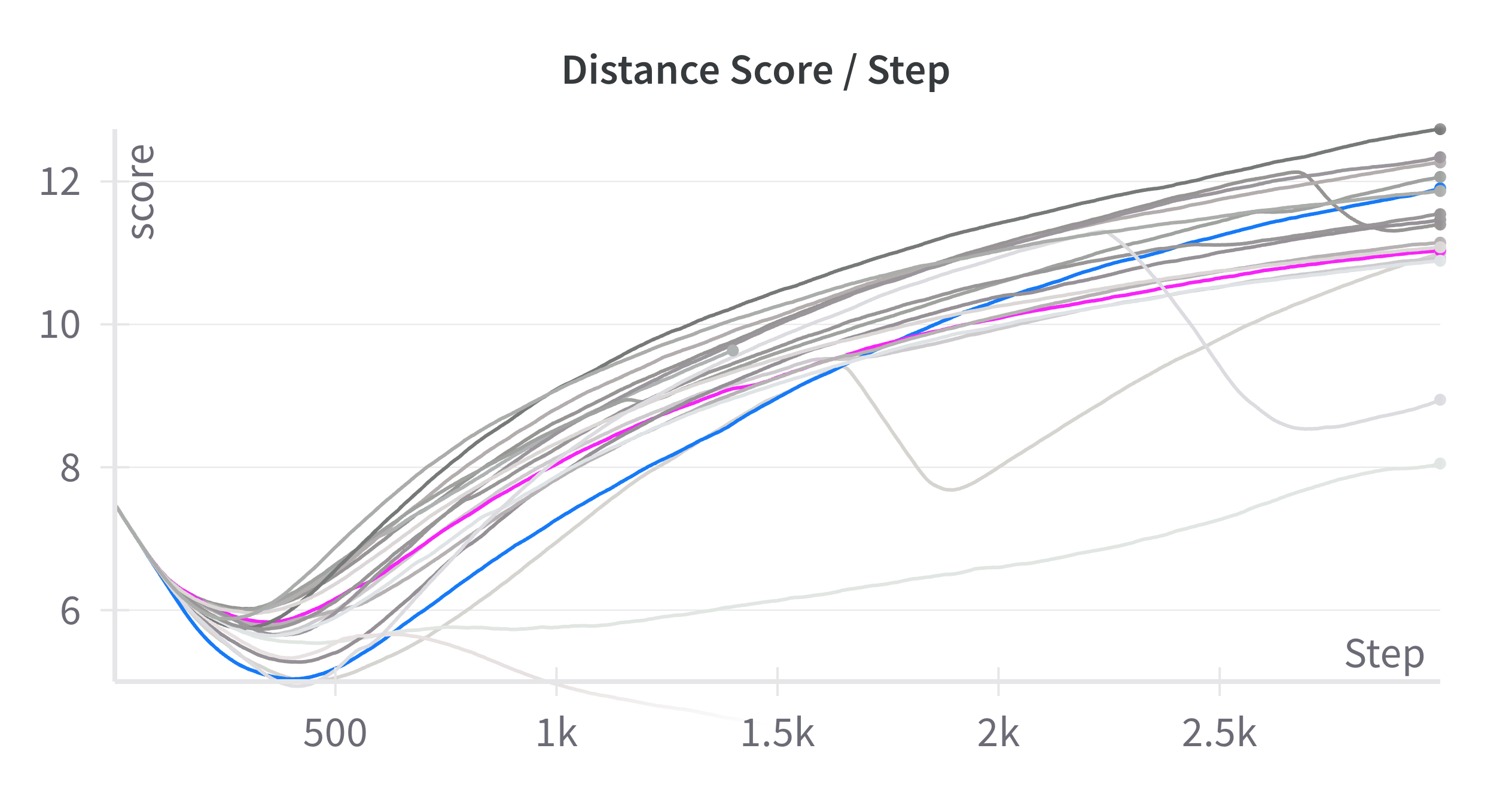}
    \caption*{}
\end{minipage}
\begin{minipage}{0.28\textwidth}
    \centering
    \includegraphics[width=\linewidth, trim={0 0 0 200}, clip]{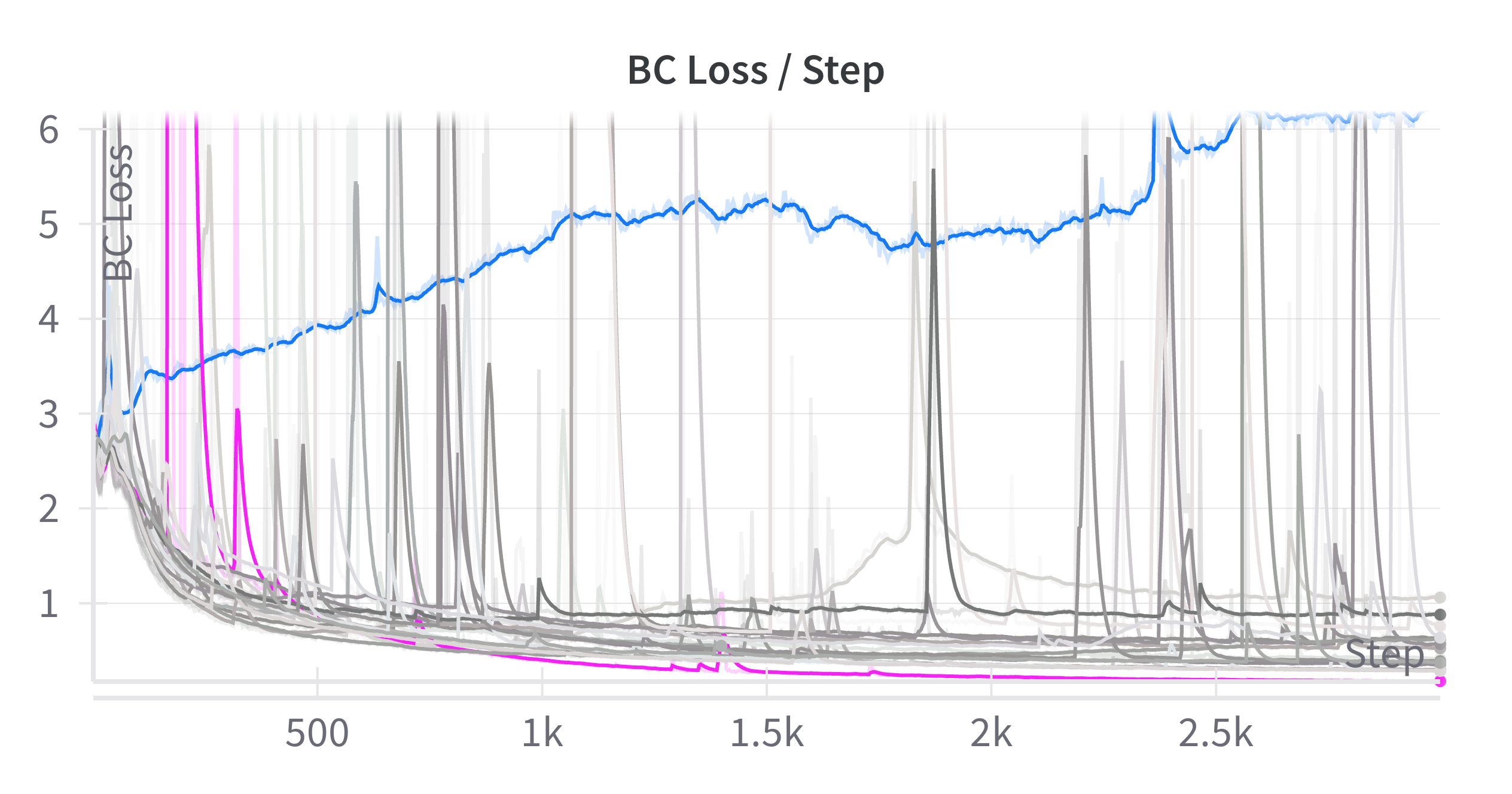}
    \caption*{}
\end{minipage}
\begin{minipage}{0.28\textwidth}
    \centering
    \includegraphics[width=\linewidth, trim={0 0 0 200}, clip]{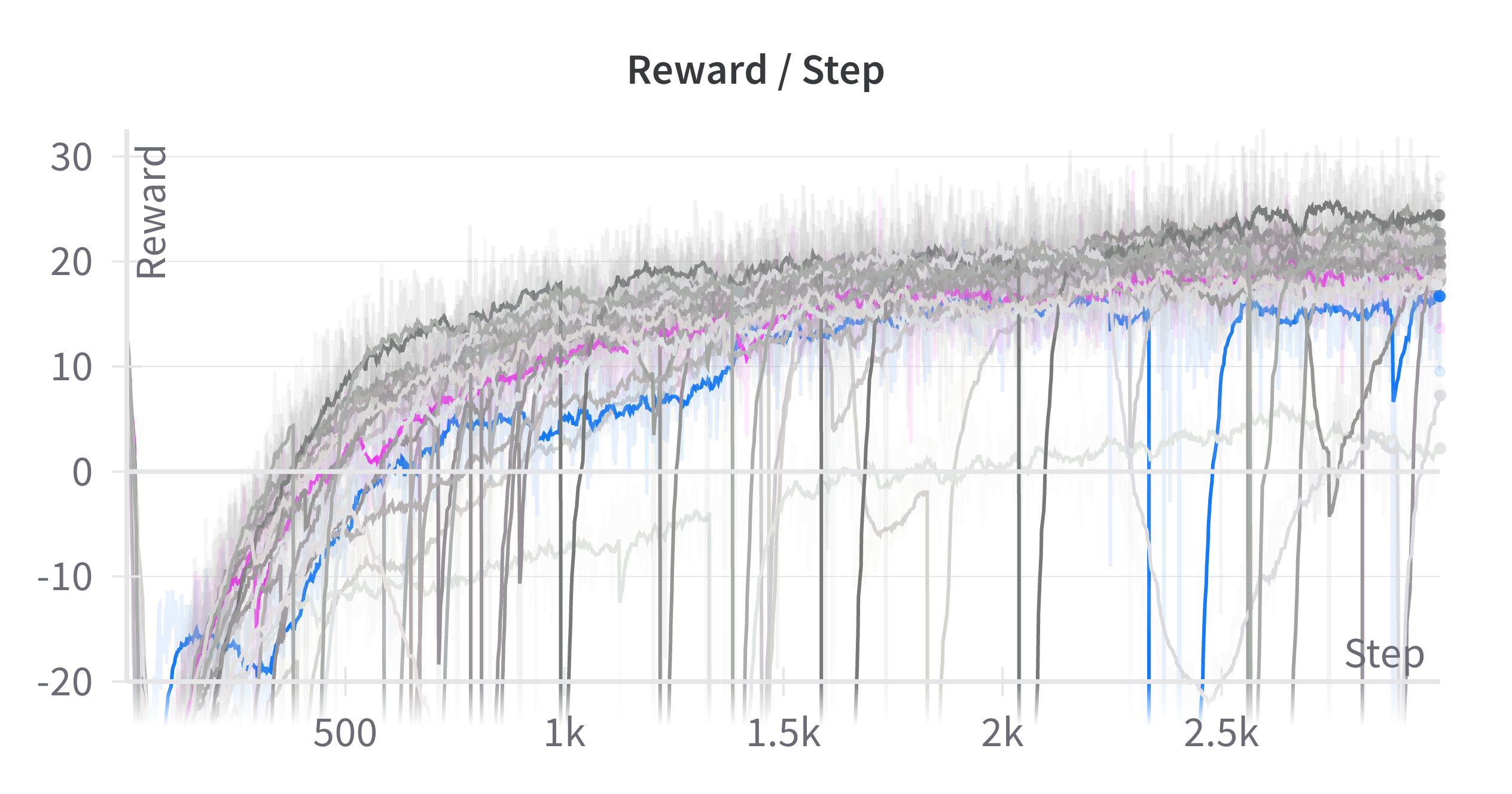}
    \caption*{}
\end{minipage}
\vspace{-0.5cm}
\caption{Comparison of 80 BC-RL transitions in which score, BC Loss, and reward plots across 3 children from 3 distinct families. The gradient from black to white ranks transitions from best to worst performance. The purple curve represents pure BC, and the blue curve represents pure RL.}

\vspace{0.1cm} 

\end{figure*}

\begin{figure*}[!htb]
\begin{center}
\vspace{-0.5cm}
\minipage{0.5\textwidth}
  \centering
  \includegraphics[width=\linewidth, trim={0 0 0 0}, clip]{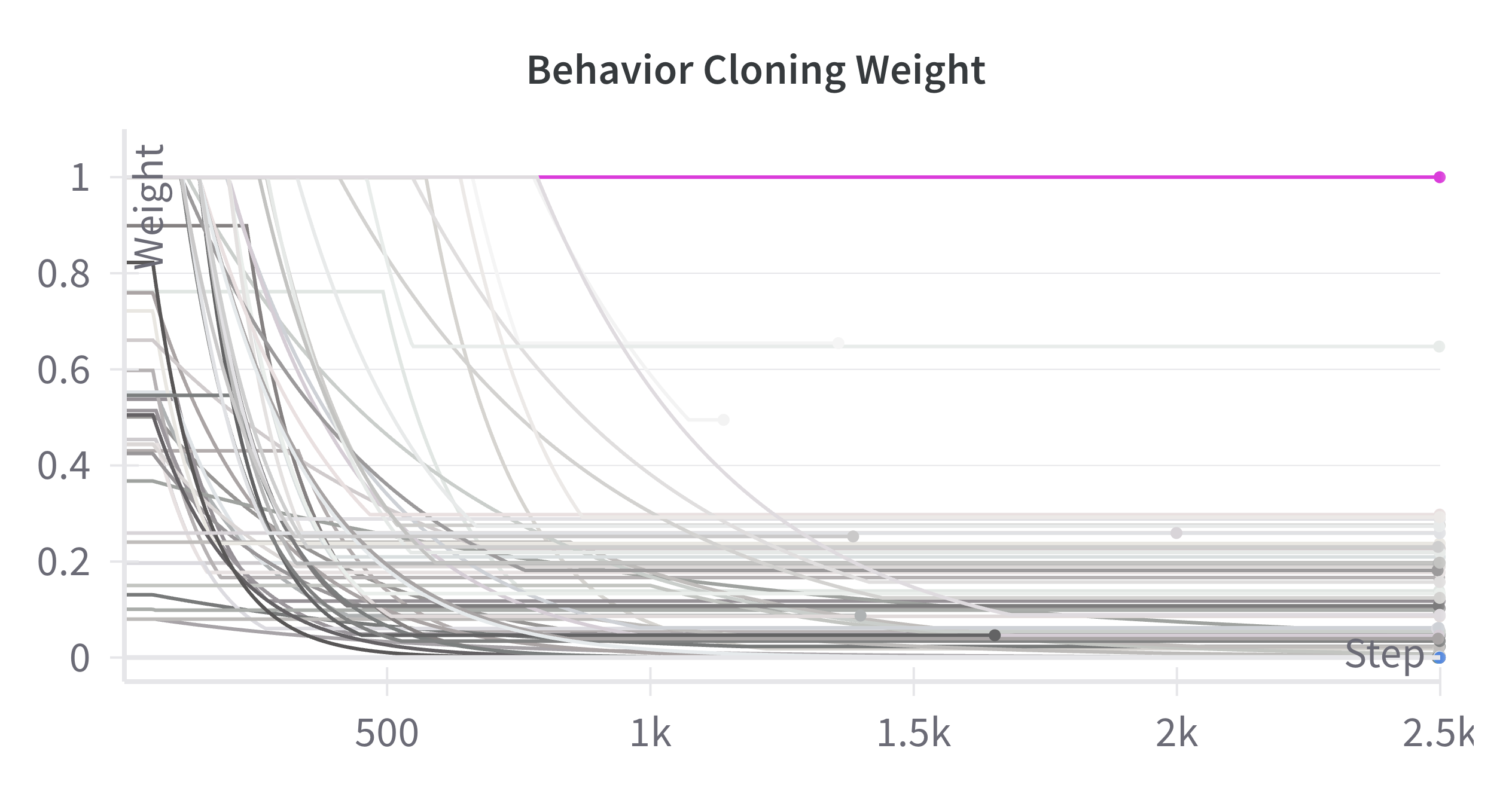}
\endminipage
\caption{Comparison relationship between BC-to-RL transitions (same data from figure 2) and their performance. Again, the gradient from black to white ranks transitions from best to worst performance. The purple curve represents pure BC, and the blue curve represents pure RL.}
\end{center}
\vspace{-0.5cm}
\end{figure*}

    We tested 80 transitions from BC to RL by progressively reducing the BC weight. The most successful transitions showed continuous improvement, leveraging parent behaviors to enhance exploration and avoid local minima. Pure BC (fixed weight at 1) demonstrated a steady reduction in BC loss, while pure RL (weight at 0) saw BC loss increase during training. Transitions that shifted to RL early, within 200 steps, maintained a BC loss belower than pure RL yet  performed better than pure BC or RL, indicating more efficient learning.

Transitions that stayed longer in the BC phase, however, remained closer to parent behaviors and failed to outperform them in later RL stages, emphasizing the need to transition to RL early enough to enable meaningful exploration and improvement.










\section{Conclusion}
\label{sec:conclusion}
We introduce an evolution-inspired optimization framework that uniquely merges the strengths of IL and RL into a reproduction module — to our knowledge, a first of its kind within an open-ended learning paradigm. The goal is to understand how transitioning from IL to RL helps agents consolidate knowledge and adapt to complex environments, where parents have specialized in different tasks. Our initial results show that this module plays a key role in efficiently merging skills from parent agents, helping offspring perform better than single-task optimization. This approach opens up opportunities for further research on how reproduction can enhance knowledge transfer and the development of more adaptable agents in the context of open-ended learning.




\nocite{*}
\bibliography{references}  


\end{document}